\documentclass[runningheads]{llncs}
\usepackage{graphicx}

\usepackage{amsmath,amssymb,amsfonts}
\begin{document}
\title{Investigating Data Memorization in 3D Latent Diffusion Models for Medical Image Synthesis}
\author{Salman Ul Hassan Dar\inst{1,2,3} \and
Arman Ghanaat\inst{1,2}  \and
Jannik Kahmann\inst{4} \and
Isabelle Ayx \inst{4}\and
Theano Papavassiliu\inst{5} \and
Stefan O. Schoenberg \inst{4} \and
Sandy Engelhardt \inst{1,2}
}

\authorrunning{Dar et al.}
\titlerunning{Memorization in 3D Latent Diffusion Models for Medical Image Synthesis}
\institute{Department of Internal Medicine III, Group Artificial Intelligence in CardiovascularMedicine, Heidelberg University Hospital, 69120 Heidelberg, Germany
\email{SalmanUlHassan.Dar@med.uni-heidelberg.de} \and
DZHK (German Centre for Cardiovascular Research), Heidelberg, Germany \and
AI Health Innovation Cluster, Heidelberg, Germany \and
Department of Radiology and Nuclear Medicine, University Medical Center Mannheim, Heidelberg University, Theodor-Kutzer-Ufer 1-3, 68167 Mannheim, Germany \and
First Department of Medicine-Cardiology, University Medical Centre Mannheim, Theodor-Kutzer-Ufer 1-3, 68167 Mannheim, Germany
}
\maketitle              % typeset the header of the contribution
\begin{abstract}
%(15-250 words)
Generative latent diffusion models have been established as state-of-the-art in data generation. One promising application is generation of realistic synthetic medical imaging data for open data sharing without compromising patient privacy. Despite the promise, the capacity of such models to memorize sensitive patient training data and synthesize samples showing high resemblance to training data samples is relatively unexplored. Here, we assess the memorization capacity of 3D latent diffusion models on photon-counting coronary computed tomography angiography and knee magnetic resonance imaging datasets. To detect potential memorization of training samples, we utilize self-supervised models based on contrastive learning. Our results suggest that such latent diffusion models indeed memorize training data, and there is a dire need for devising strategies to mitigate memorization. 
\keywords{Deep generative models \and Latent diffusion \and Data memorization \and Patient privacy \and Contrastive learning}
\end{abstract}
\section{Introduction}
\begin{figure}[t]
\centering
\includegraphics[width=0.9\textwidth]{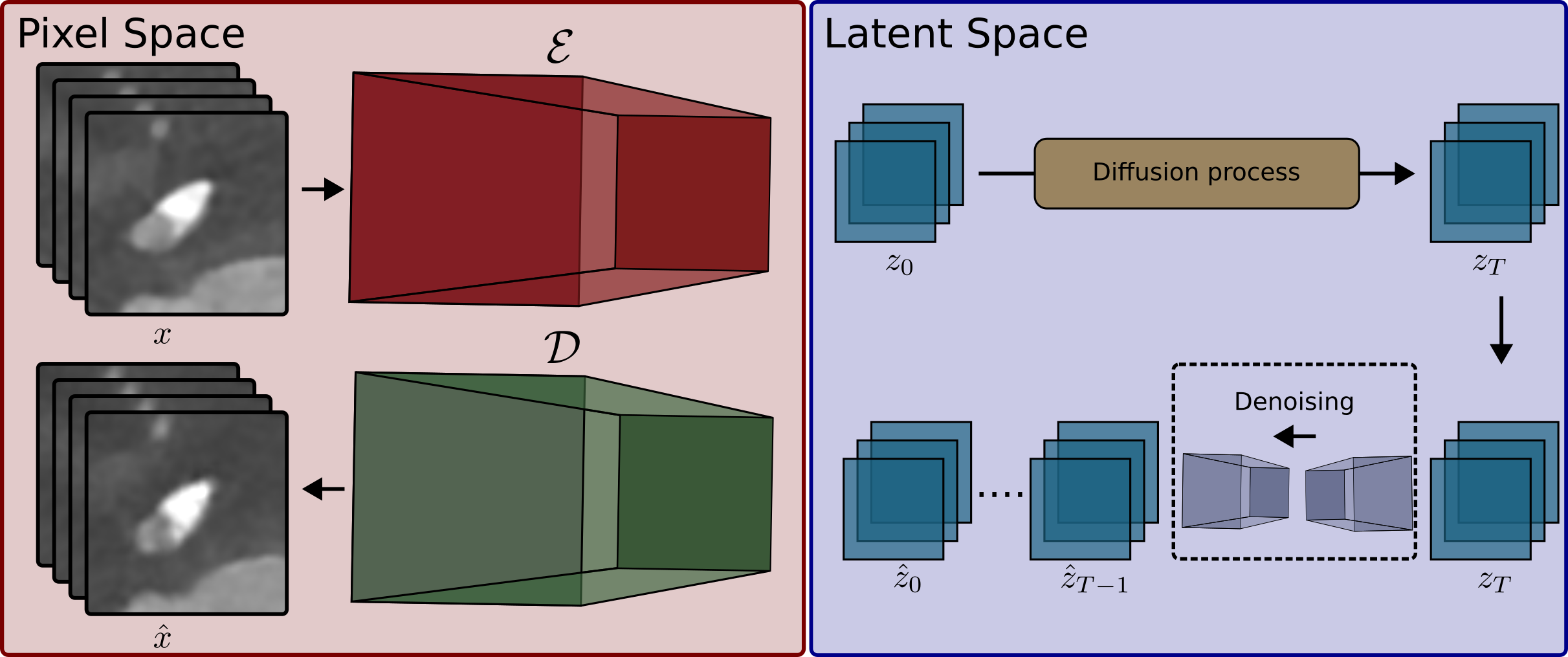}
\caption{3D Latent diffusion models first project 3D sub-volumes onto a lower dimensional latent space for computational efficiency using an encoder. Diffusion models are then trained to gradually denoise the noisy latent space representation. Upon complete denoising, the representation is projected back onto the pixel space using a decoder.} \label{diffusion}
\end{figure}

Contemporary developments in deep generative modeling have lead to performance leaps in a broad range of medical imaging applications \cite{Ganreview2019,diffusion4medreview,diffusionanomaly,diffusionrecon,diffusionsegmentation,diffusionimagetranslation}. One promising application is generation of novel synthetic images \cite{Khader2023,Pinaya2022,dorjsembe2022threedimensional,sandymiccai,generativelaproscopic}. Synthetic images can be used for data diversification by synthesizing samples belonging to underrepresented classes for training of data-driven models or sharing of synthetic data for open science without compromising patient privacy. \\
State-of-the art generative models are based on latent diffusion \cite{Rombach2022a}. These models first project data onto a compressed latent space, learn latent space distribution through a gradual denoising process, and synthesize novel latent space samples followed by projection onto a high dimensional pixel space \cite{Rombach2022a}. Despite the ability to synthesize high quality samples, recent studies in computer vision suggest that latent diffusion models (LDMs) are prone to training data memorization \cite{Somepalli2023,Somepalli2023understanding,Carlini2023extracting}. This can be more critical in medical imaging, where synthesizing real patient data defeats the whole purpose of preserving data privacy. These computer vision studies further suggest that the phenomenon of data memorization is more prevalent in low data regimes \cite{Somepalli2023}, which is very often the case in the medical domain. Despite the importance of patient privacy, it is surprising that data memorization in generative models has received little attention in the medical imaging community.\\
Here, we investigate the memorization capacity of 3D-LDMs in medical images. To this end, we train LDMs to generate 3D volumes (Fig. \ref{diffusion}) and compare novel generated samples with real training samples via self-supervised models (Fig. \ref{contrastive_learning}) for potential memorization. For assessment, we perform experiments on an in-house photon-counting coronary computed angiography (PCCTA) dataset and a public knee MRI (MRNet) dataset \cite{MRNet}. Our results suggest that LDMs indeed suffer from training data memorization. 
\subsection{Data Generation via LDMs}
LDMs belong to a family of generative models that learn to generate novel realistic samples by denoising normally distributed noise in a compressed lower dimensional latent space \cite{Rombach2022a}. LDMs consist of two models:
\subsubsection{Latent Encoding Model}
First, an encoder learns to project samples onto lower dimensional latent space. This lower dimensional latent space is typically learned using an autoencoder. The autoencoder is trained to encode the image $x \in \mathbb{R}^{L \times H \times W}$ to a latent space $z \in \mathbb{R}^{L' \times H' \times W'}$ using an encoder $\mathcal{E}$ having parameters $\theta_{\mathcal{E}}$ ($z = \mathcal{E}(x) $), followed by reconstruction via a decoder $\mathcal{D}$ having parameters $\theta_{\mathcal{D}}$ ($\hat{x} = \mathcal{D}(z)$).
% \begin{align}
% \label{eq:autoencoder}
% z = \mathcal{E}(x) \notag \\
% \hat{x} = \mathcal{D}(z)
% \end{align}
Overall, the training is performed to minimize the following reconstruction loss function:
\begin{align}
\label{eq:autoencoder_loss}
\mathcal{L}_{rec}(\theta_{\mathcal{E}}, \theta_{\mathcal{D}}) = \mathbb{E}_{p(x)}\left[\left\|x - \hat{x} \right\|_{1}\right]
\end{align}
where $\mathbb{E}_{p(x)}$ denotes expectation with respect to data distribution $p(x)$. Since the encoder and decoder are trained simultaneously with the aim to recover the original image from a lower dimensional representation, the encoder learns to project the data onto semantically meaningful latent space without loosing much information.
\subsubsection{Diffusion Model}
Afterwards, a deep diffusion probablistic model (DDPM) is trained to recover meaningful latent space from normally distributed noise. DDPMs consist of a forward and reverse diffusion step. In the forward step normally distributed noise is added to the latent representation ($z$) in small increments. 
At any time $t$, the relation between $z_t$ and $z_{t-1}$ can be expressed as:
\begin{eqnarray}
\label{eq:diff_forward}
q\left( z_{t}|z_{t-1} \right) &=&\mathcal{N}\left( x_{t}; \sqrt{1-\beta_{t}}z_{t-1},\beta_{t}\mathrm{I} \right)
\end{eqnarray}
where $\beta_{t}$ is the variance schedule \cite{DDPM} and $q\left( z_{t}|z_{t-1} \right)$ is the conditional distribution. In the reverse step, a model is trained to approximate $q\left( z_{t-1}|z_{t} \right)$ as $p_{\theta}\left( z_{t-1}|z_{t} \right)$.
Once trained, the model can be used to synthesise novel representations ($z_0$) given $z_{T}\sim \mathcal{N}\left( 0,\mathrm{I} \right)$. The latent code $z_0$ can then be fed as input to the decoder ($\mathcal{D}$) to generate novel samples from data distribution $p(x)$.

\begin{figure}[t]
\centering
\includegraphics[width=0.9\textwidth]{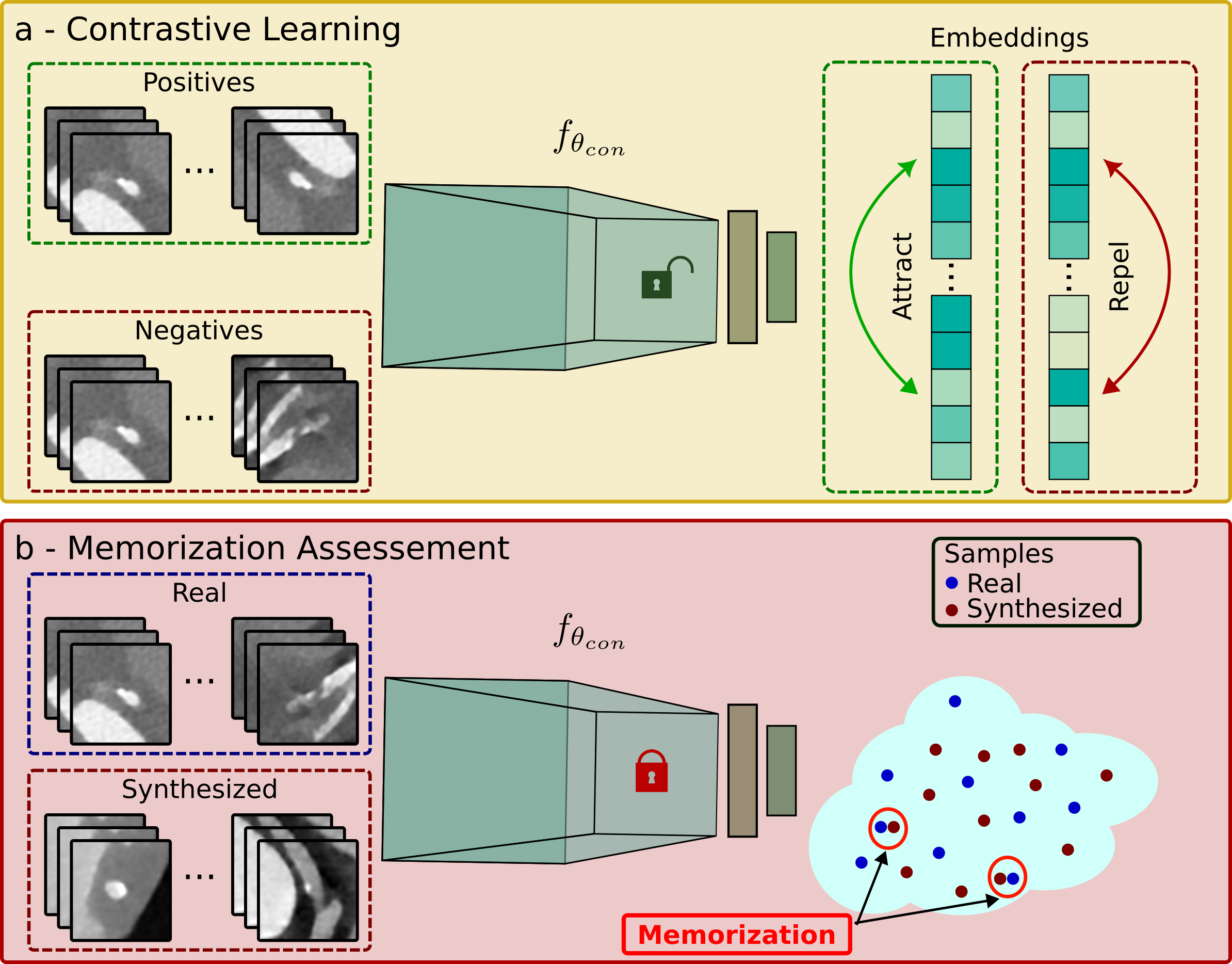}
\caption{ \textbf{a} - A self-supervised model is trained based on contrastive learning to learn a lower dimensional embedding where augmented versions of the same sample are attracted and different samples are repelled. \textbf{b} - The trained model is then used to identify if the synthetic samples are copies of the real samples} \label{contrastive_learning}
\end{figure}
\subsection{Memorization Assessment}
Although LDMs have outperformed their counterpart generative models in medical image synthesis in terms of image quality and diversity \cite{Khader2023,Pinaya2022}, their capacity to memorize training samples remains relatively unexplored. This is surprising, considering that one of the main goals of sharing synthetic data is to preserve patient privacy. Memorization of patient data defeats this purpose, and the quality of the synthesized samples becomes secondary. \\
Since the primary focus of this work is memorization,  it is important to first define what constitutes ``memorization''. Akbar et al. \cite{akbar2023beware} define memorization as a phenomenon where generative models can generate copies of training data samples. However, they do not explicitly define what a ``copy'' means, and their use of ``copy'' seems to be limited to a synthesised sample that is identically oriented to a training sample and shares the same anatomical structures with minor differences such as image quality. They detect potential copy candidates by computing pairwise normalized correlation between synthesized and training samples. 
This fails to take into account that synthetic samples can also be flipped or rotated versions of the training samples, which can easily result as a consequence of data augmentation strategies typically used for training deep models.
In our work, we expand the definition of ``copy'' to further include rotated and flipped versions of a training sample. This definition can further be broadened to include other forms of variations such as slight deformation. However, for simplicity here we limit the additional variations to flipping and rotation.
To detect potential copies, we first train a self-supervised model based on the contrastive learning approach (Fig. \ref{contrastive_learning}). The aim is to have a low dimensional latent representation of each sample such that the augmented versions of a sample have similar latent representations, and different samples have distinct representations. 
The model is trained to minimize the following loss function \cite{tripletloss}:
\begin{align}
\label{eq:autoencoder}
\mathcal{L}_{con}(\theta_{con}) = \mathbb{E}_{p(x)}\left[ \max(0, \left\|f_{\theta_{con}}(x) - f_{\theta_{con}}(x^{+}) \right\|_{2} - \left\|f_{\theta_{con}}(x) - f_{\theta_{con}}(x^{-}) \right\|_{2}) \right]
\end{align}
where $x$ corresponds to a training volume, $x^{+}$ is the similar sample which is an augmented version of $x$, $x^{-}$ denotes the dissimilar sample which is just a different volume, and $f_{\theta_{con}}(.)$ is the networks with trainable parameters $\theta_{con}$ that maps input $x$ to a low dimensional representation.\\
After training $f_{\theta_{con}}(.)$, we compare the embeddings of the synthesized samples with the real samples in the low dimensional representational space.
\section{Methods}
\subsection{Datasets}
To demonstrate memorization in medical imaging, we selected two datasets covering a range of properties in terms of imaging modalities, organs, resolutions, 3D volume sizes, and dataset sizes. We conducted experiments on in house photon-counting coronary computed tomography angiography (PCCTA) dataset and a publicly available knee MRI (MRNet) dataset \cite{MRNet}. PCCTA images were acquired from 65 patients on a Siemens Naeotom Alpha scanner at the xxx xxx xxx. Ethics approval was approved by the ethics committee of xxx (ID xxx). Images were acquired with a resolution of approximately 0.39mm$\times$0.39mm$\times$0.42mm. In all patients, coronary artery plaques were annotated by an expert radiologist. 
Sub-volumes of size 64$\times$64$\times$64 surrounding plaques were extracted, resulting in 242 sub-volumes for training and 58 sub-volumes for validation in total.  In MRNet, T2-weighted knee MR images of 1130 subjects were analyzed, where 904 subjects  were used for training and 226 for validation. All volumes were cropped or zero-padded to have sizes of 256$\times$256$\times$32. In both datasets, each volume was normalized to have voxel intensity in the range [-1, 1].
\subsection{Networks}
LDM architecture, training procedures and loss functions were directly adopted from Khader et al. \cite{Khader2023}. For the training of the diffusion and autoencoder models, all hyperparameters  were matched with the ones selected in Khader et al. \cite{Khader2023}. The only exception was the batch size in the diffusion models, which was set to 10 to fit models into the GPU VRAM. 
For contrastive learning, network architecture was adopted from the encoder in the latent encoding model. The encoder was used to reduce the sub-volume dimensions to  4$\times$4$\times$4 and 8 channels. Afterwards, flattening was performed followed by two densely connected layers to reduce the latent space embeddings to dimensions 32$\times$1. All hyperparameters except for the learning rate and epochs were identical to the latent encoding model. Learning rate and epochs were tuned using a held out validation data. 
\section{Results}
\subsection{Memorization Assessment}
\begin{figure}[t]
\centering
\includegraphics[width=0.9\textwidth]{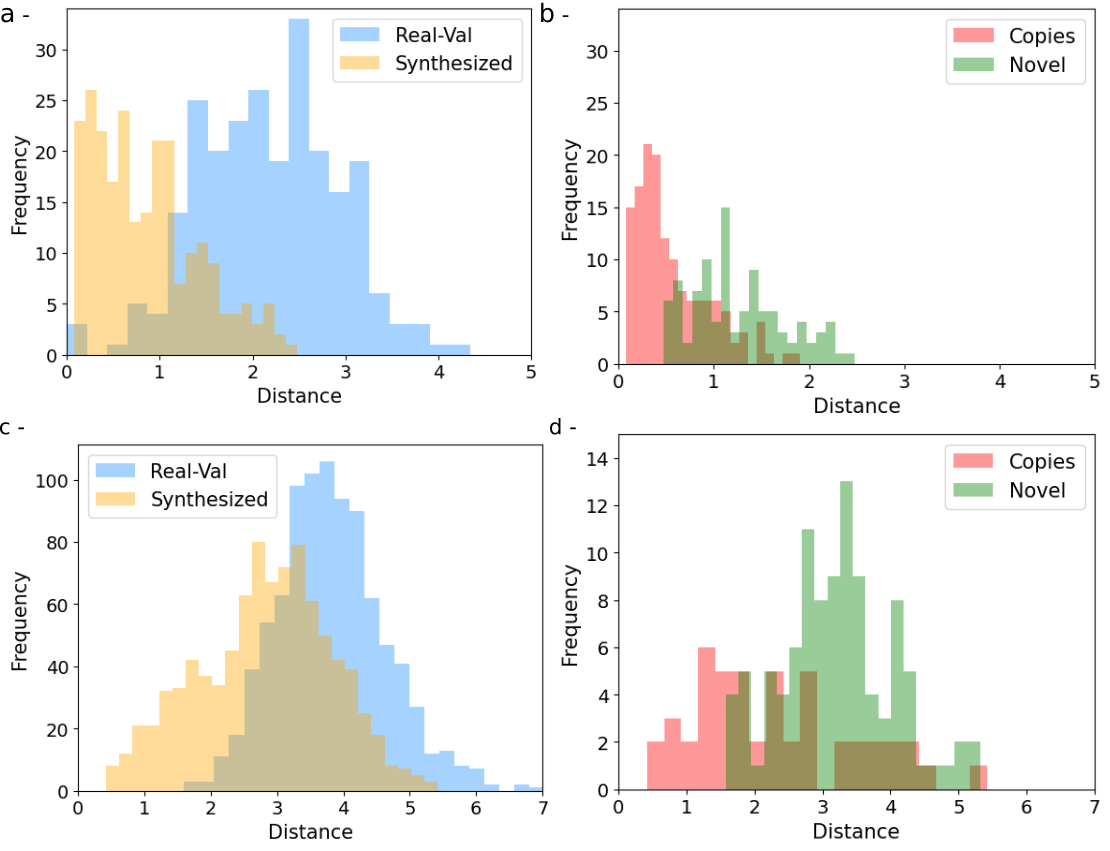}
\caption{MSD distributions of copy candidates (Synthesized) and closest validation samples (Real-Val) in \textbf{a} -  PCCTA and \textbf{c} -  MRNet datasets are shown. Higher density near zero implies more similarity. MSD distributions of copy candidates in \textbf{b} - PCCTA and  \textbf{d} - MRNet datasets manually annotated as copies or novel samples are shown. } \label{histograms1}
\end{figure}

\begin{figure}[t]
\centering
\includegraphics[width=1\textwidth]{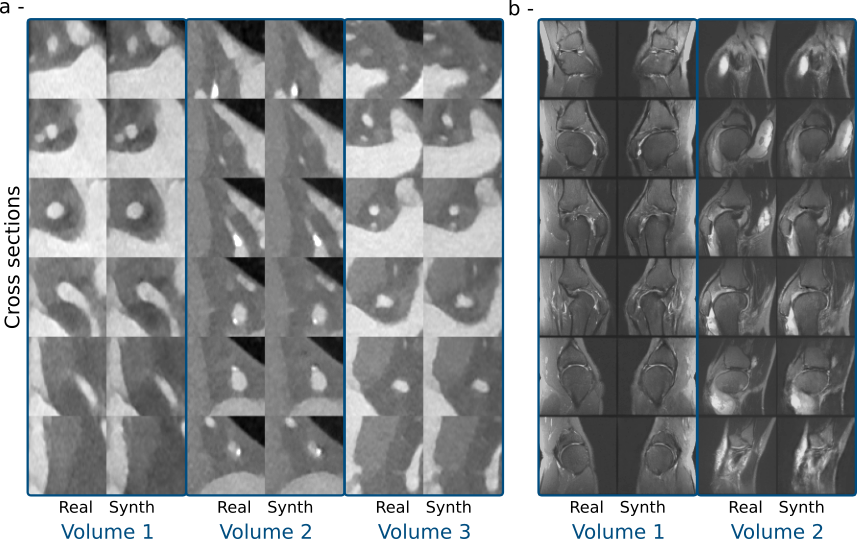}
\caption{Representative samples and copy candidates from the \textbf{a} - PCCTA and \textbf{b} - MRNet dataset are shown. Columns corresponds to real or synthesized (Synth) samples, and rows correspond to six cross sections selected from the sub-volumes.  Synthesized samples have a stark resemblance with the real training samples. A copy candidate that is a flipped version of a training sample is also shown (b - Volume 1).} \label{memorized_samples}
\end{figure}

\begin{figure}
\centering
\includegraphics[width=0.9\textwidth]{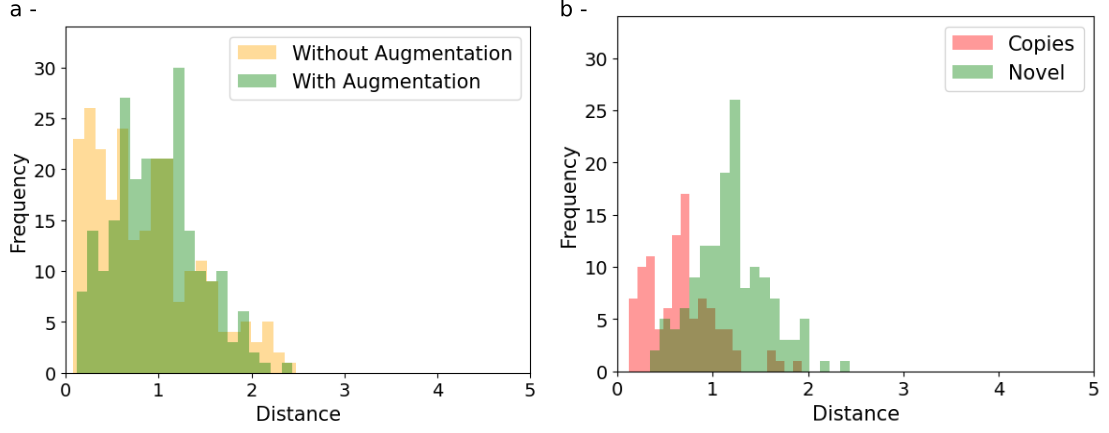}
\caption{\textbf{a} - MDS distribution of copy candidates for models trained with and without data augmentation. \textbf{b} - MSD distributions of copy candidates manually annotated as copies or novel samples on models trained with data augmentation. } \label{histograms2}
\end{figure}

First, 1000 synthetic PCCTA samples (approx. 4 $\times$ training data) were generated using the trained LDM. All synthetic and training samples were then passed through the self-supervised models (Section 1.2) to obtain corresponding lower dimensional embeddings. Next, mean square distance (MSD) was computed between all training and synthetic embeddings. For each training sample, the closest synthetic sample was considered as a copy candidate. Fig. \ref{histograms1}a shows MSD distribution of the candidate copies.
To get a better idea of the MSD scale, for each training sample the closest real validation sample in the embedding space was also considered (Fig. \ref{histograms1}a). 
Low values on the x-axis denote lower distance or high similarity.
%If the difference was smaller than $\epsilon$ the sample was considered as a copy.
The MSD distribution of synthetic samples is more concentrated near zero compared to the MSD distribution of real validation samples.\\ 
To further assess if the candidate synthesized samples are indeed copies, each candidate copy was also labelled manually as a copy or a novel synthetic sample via visual assessment by consensus of  two users. As shown in Fig. \ref{histograms1}b, most of the candidates with low MSD values are copies.
Upon comparing copies with novel samples in \ref{histograms1}b, we also observe that 59\% of the training data has been memorized. This number is alarming, as it indicates memorization at a large scale. It is also important to note that this percentage is based on just 1000 synthesized samples. Increasing the synthetic samples could lead to an increased number of copies. Fig. \ref{memorized_samples}a shows some copy candidates. It can be seen that synthetic samples show stark resemblance with the training samples. \\
We then assess memorization in the MRNet dataset, which is relatively a larger dataset containing 904 training volumes. 3600 synthetic samples (approx. 4 $\times$ training data) were generated using the LDM trained on the MRNet dataset. Fig. \ref{histograms2}c shows MSD distribution of synthetic candidate copies and validation samples. We observe similar patterns in the MRNet dataset. However, MSD distribution of synthetic candidate copies in MRNet is further away from zero compared to the PCCTA dataset. This can be explained by the training data size, as models with large training datasets get to learn distribution from many diverse samples and thus are less likely to memorize the data. We also annotated 150 randomly selected copy candidates as copy or novel samples (Fig. \ref{histograms2}d). We find 33\% of the copy candidates to be copies. Fig. \ref{memorized_samples}-b also shows representative samples. 
\subsection{Data Augmentation}
We also analyze the effect of data augmentation on memorization by comparing MSD distribution of copy candidates generated by models trained with augmentation (augmented models) and without augmentation (non-augmented models) on the PCCTA dataset. Fig. \ref{histograms2} compares the MSD distributions. MSD distribution of the non-augmented model tends to have higher density near zero. Overall, 41\% of the training dataset is memorized in augmented models compared to 59\% in non-augmented models. This suggests that the non-augmented models tend to memorize more than the augmented models. One possible explanation could be artificial expansion of datasets  through augmentation, which can in turn lead to fewer repetitions of identical forms of a sample during training.
\section{Discussion}
There has been a considerable amount of focus on generative models in medical image synthesis. 
Here, we tried to assess if these models actually learn to synthesize novel samples as opposed to memorizing training samples. Our results suggest that LDMs indeed memorize samples from the training data. This can have broad implications in the medical imaging community, since leakage of patient data in the form of medical images can lead to violation of patient privacy. \\
An interesting future prospect could be to understand the underlying reasons leading to memorization. Somepalli et al. \cite{Somepalli2023understanding} suggests that data duplication during training could be an important factor, as a repeated sample is seen many more times during training. 
They further suggest that unconditional models primarily suffer from data memorization in low data regimes, which is similar to what we observe when we compare memorization in PCCTA and MRNet datasets. Nonetheless, it is an important research direction which warrants future work.\\
To our knowledge, this is the first  study assessing memorization in 3D LDMs for medical image synthesis. Another independent study recently investigated memorization in deep diffusion models \cite{akbar2023beware}. There are several differences between our study and Akbar et al. : 1) Akbar et al. trained 2D models on medical images, whereas we trained 3D models which are more coherent with the nature of the medical images. 2) Akbar et al. is based on diffusion models in pixel space, which are not easily applicable to 3D medical images due to high computational demands. To the contrary, here we used LDMs, which first project the data onto a low dimension latent space to reduce computational complexity while ensuring that the relevant semantic information is preserved. 3) Akbar et al. used correlation between images in the pixel space to assess memorization. While this approach detects identical copies, it cannot detect augmented or slightly different copies. Here, we trained a self-supervised model that can also account for augmented versions of the training samples, which might be missed by computing regular correlations in pixel space. 4) We also assessed memorization on models that were trained using augmented data. This is a more realistic scenario while training deep models on medical images due to data scarcity.
\subsubsection{Acknowledgment:} This work supported through state funds approved by the State Parliament of Baden-Württemberg for the Innovation Campus Health + Life Science Alliance Heidelberg Mannheim.
In addition, this work was supported by the BMBF-SWAG Project 01KD2215D.
The authors also gratefully acknowledge the data storage service SDS@hd supported by the Ministry of Science, Research and the Arts  Baden-Württemberg (MWK) and the German Research Foundation (DFG) through grant INST 35/1314-1 FUGG and INST 35/1503-1 FUGG.

\bibliographystyle{splncs04}
\bibliography{Papers}

\end{document}